\documentclass[sigconf]{aamas}

\usepackage{balance} 

\usepackage{amsmath}

\usepackage{amssymb}
\usepackage{mathtools}
\usepackage{amsthm}
\usepackage{bbm}
\usepackage[capitalize, noabbrev]{cleveref}
\usepackage{graphicx}
\usepackage{wrapfig}
\usepackage{makecell}
\usepackage{enumitem}
\usepackage{algorithm}
\usepackage{algorithmic}
\usepackage{tcolorbox}
\usepackage{caption}
\newcommand{\Xcal}{\mathcal{X}}
\newcommand{\Scal}{\mathcal{S}}
\newcommand{\Acal}{\mathcal{A}}

\newcommand{\Ucal}{\mathcal{U}}
\newcommand{\Ocal}{\mathcal{O}}

\newtheorem{problem}{Problem}
 
\DeclareMathOperator{\subjectto}{s.t.}

\setcopyright{ifaamas}
\acmConference[AAMAS '26]{Proc.\@ of the 25th International Conference
on Autonomous Agents and Multiagent Systems (AAMAS 2026)}{May 25 -- 29, 2026}
{Paphos, Cyprus}{C.~Amato, L.~Dennis, V.~Mascardi, J.~Thangarajah (eds.)}
\copyrightyear{2026}
\acmYear{2026}
\acmDOI{}
\acmPrice{}
\acmISBN{}

\acmSubmissionID{1326}

\title[AAMAS-2026 CoNav-Maze]{IG-MCTS: Human-in-the-Loop Cooperative Navigation under Incomplete Information}

\author{Shenghui Chen}
\authornote{Equal contribution.}
\affiliation{
  \institution{The University of Texas at Austin}
  \city{Austin}
  \state{Texas}
  \country{USA}}
\email{shenghui.chen@utexas.edu}

\author{Ruihan Zhao}
\authornotemark[1]
\affiliation{
  \institution{The University of Texas at Austin}
  \city{Austin}
  \state{Texas}
  \country{USA}}
\email{ruihan.zhao@utexas.edu}

\author{Sandeep Chichali}
\affiliation{
  \institution{The University of Texas at Austin}
  \city{Austin}
  \state{Texas}
  \country{USA}}
\email{sandeepc@utexas.edu}

\author{Ufuk Topcu}
\affiliation{
  \institution{The University of Texas at Austin}
  \city{Austin}
  \state{Texas}
  \country{USA}}
\email{utopcu@utexas.edu}

\begin{abstract}
    Human-robot cooperative navigation is challenging under incomplete information. We introduce CoNav-Maze, a simulated environment where a robot navigates with local perception while a human operator provides guidance based on an inaccurate map. The robot can share its onboard camera views to help the operator refine their understanding of the environment. To enable efficient cooperation, we propose Information Gain Monte Carlo Tree Search (IG-MCTS), an online planning algorithm that jointly optimizes autonomous movement and informative communication. IG-MCTS leverages a learned Neural Human Perception Model (NHPM)---trained on a crowdsourced mapping dataset---to predict how the human’s internal map evolves as new observations are shared. User studies show that IG-MCTS significantly reduces communication demands and yields eye-tracking metrics indicative of lower cognitive load, while maintaining task performance comparable to teleoperation and instruction-following baselines. 
    Finally, we illustrate generalization beyond discrete mazes through a continuous-space waterway navigation setting, in which NHPM benefits from deeper encoder–decoder architectures and IG-MCTS leverages a dynamically constructed Voronoi-partitioned traversability graph.
\end{abstract}


\newcommand{\BibTeX}{\rm B\kern-.05em{\sc i\kern-.025em b}\kern-.08em\TeX}

\begin{document}


\pagestyle{fancy}
\fancyhead{}


\maketitle 


\section{Introduction}\label{sec:intro}
Effective collaboration between humans and robots in complex environments requires efficient information exchange under uncertainty. In many real-world scenarios—such as search and rescue, exploration, or remote inspection—robots operate in partially observed spaces where human operators possess complementary task understanding but face limited communication bandwidth. This asymmetry poses a fundamental coordination challenge: how can an autonomous agent selectively communicate or act to maximize shared situational awareness and task performance without continuous supervision?

Current human-in-the-loop robot control approaches differ in the degree of control required and the communication bandwidth utilized. Teleoperation relies on low-level human inputs, leading to high cognitive load and communication demands~\cite{moniruzzaman2022teleoperation}. Instruction-following, in contrast, reduces direct human control by allowing robots to execute high-level human plans \cite{anderson2018vision}. However, this approach still requires substantial communication for the human to provide detailed and accurate instructions. These limitations highlight the need for a more autonomous approach that reduces human workload while maintaining effective collaboration.

\begin{figure*}[t]
    \centering
    \includegraphics[width=\linewidth]{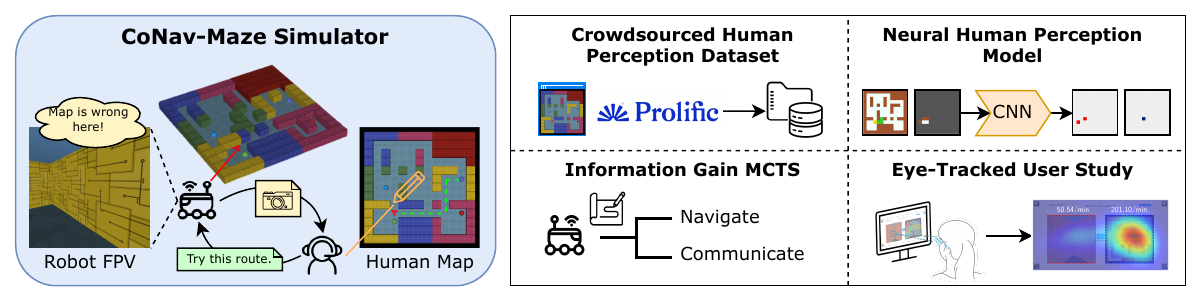}
    \caption{We enable efficient human-robot collaboration in a maze navigation setting. \textbf{Left:} The robot gathers local observations, while the human navigates using an imprecise global map. The robot can transmit images to improve the human’s understanding of the environment, while the human assists by suggesting paths. \textbf{Right:} Main contributions: (1) crowdsourcing a dataset of human perceptual updates, (2) training a neural human perception model, (3) developing Information Gain Monte Carlo Tree Search (IG-MCTS), a planning algorithm that balances task progress with informative communication. (4) validating the approach through a user study with eye-tracking and task metrics.}
    \Description{Diagram of human-robot maze navigation task and a summary of contributions.}
    \label{fig:overview_figure}
    \vspace{-1em}
\end{figure*}

This paper studies the human-robot cooperative navigation problem in a simulated environment called \emph{CoNav-Maze}, where the robot obtains local SLAM observations while the human provides guidance based on an initially inaccurate global map. The goal is to develop a robot control algorithm that does not merely follow instructions but actively collaborates by transmitting images to improve the human's understanding of the environment, integrating human trajectory guidance, and maintaining sufficient autonomy to reach target locations efficiently.

We introduce \emph{Information Gain Monte Carlo Tree Search} (IG-MCTS), an online planning algorithm that embodies the idea that \emph{communication is action}. Aside from task-centric objectives, IG-MCTS strategically decides between movement and communication actions based on their potential to enhance the human’s understanding of the environment.
Inspired by evidence that humans read to minimize perceptual errors and extract relevant features under limited processing capacity~\cite{kangassalo2020information}, we hypothesize that a similar cognitive strategy applies to visual tasks.
To align with this cognitive pattern, IG-MCTS chooses camera angles that maximize an information reward that measures the change in the human's perception of the environment.
IG-MCTS also incorporates human-guided trajectories as reward augmentations~\cite{chen2025human}.

At the core of IG-MCTS is the \textit{Neural Human Perception Model (NHPM)}, a data-driven model that predicts how the human’s internal map evolves in response to the robot’s movements and shared observations. NHPM is implemented as a fully convolutional neural network that integrates spatial context and captures how human perception changes when new visual information is communicated by the robot. To train this model, we crowdsource a dataset of human–robot mapping interactions in \emph{CoNav-Maze}, where participants iteratively update their maps based on the robot’s camera images and trajectories. This dataset reveals characteristic human information-processing patterns during cooperative navigation. The resulting NHPM effectively learns to approximate these perceptual dynamics, achieving higher prediction accuracy than a psychometric function-based baseline~\cite{treutwein1999fitting}.

We evaluate the performance of IG-MCTS in CoNav-Maze against two baselines: teleoperation and instruction-following. 
A user study with $14$ eye-tracked participants shows that interacting with the IG-MCTS robot significantly reduces communication demands and yields eye-tracking metrics indicative of lower cognitive load, all while maintaining task performance on par with the baselines.

Finally, to demonstrate the scalability and generality of the proposed framework, we extend both NHPM and IG-MCTS beyond discrete mazes to continuous environments. In an illustrative simulated waterway navigation example, NHPM scales up using a deeper encoder--decoder architecture capable of modeling perception updates over high resolution, smooth traversable regions and obstacles, while IG-MCTS operates on a dynamically constructed graph with Voronoi partitioning to support planning over continuous space. This extension highlights that the proposed approach is not limited to grid-based domains but can generalize to more realistic navigation settings that capture the complexity of real-world human-robot cooperation.

\section{Related Work}\label{sec:related}

\paragraph{Human Perception Modeling.}
Human perception has been extensively studied in psychophysics~\cite{prins2016psychophysics}, which examines how physical stimuli influence human sensations and perceptions.
A key model is the psychometric functions, which characterize the probabilistic relationship between stimulus intensity and observer responses~\cite{wichmann2001psychometric,klein2001measuring}. The logistic psychometric function is a widely used variant that captures the gradual transition from detecting to not detecting a stimulus as intensity changes~\cite{treutwein1999fitting}. While psychometric functions provide a structured framework for modeling perception, their reliance on predefined mathematical forms may limit their ability to capture more nuanced perceptual phenomena~\cite{morgan2012observers,garcia2013shifts}.

\paragraph{Teleoperation.}
Teleoperation allows human operators to remotely control robots, facilitating tasks in hazardous or inaccessible environments. Advances in immersive interfaces and reduced latency have improved human-robot interaction~\cite{moniruzzaman2022teleoperation}, but challenges remain, including reliance on stable communication links and maintaining operator situational awareness under high cognitive load. The cognitive burden required for direct control in teleoperation often limits the human-to-robot ratio~\cite{murphy2004human}, as it becomes challenging to effectively manage multiple robots simultaneously.

\paragraph{Vision-and-Language Navigation (VLN).}
VLN tasks require agents to interpret human instructions and navigate through 3D environments by executing action sequences~\cite{anderson2018vision}. 
Methods have leveraged synthetic data generation, cross-modal imitation learning, and reinforcement learning with look-ahead planning~\cite{fried2018speaker,wang2019reinforced,wang2018look}. 
However, many existing methods rely on simplifying assumptions, such as complete information, static environments~\cite{kolve2017ai2,puig2023habitat,xia2018gibson}, or panoramic action spaces~\cite{fried2018speaker}. Recently, KnowNo~\cite{ren2023robots} explores the use of conformal prediction to quantify uncertainty in robotic scene recognition, but it restricts human input to predefined multiple-choice options.
In contrast, we focus on human-robot coordination in incomplete and dynamic environments, allowing users to specify new routes that evolve as new information is gathered.

\section{Human-Robot Cooperative Navigation Problem}\label{sec:problem}
We study a human-robot cooperative navigation task under incomplete information. The human operator has an outdated global map, while the robot can acquire accurate local observations. The human provides high-level guidance, and the robot shares local visual observations. They jointly aim to reach goal locations efficiently. We introduce \emph{CoNav-Maze}, a simulated maze environment adapted from MemoryMaze~\cite{pasukonis2022evaluating}. In this environment, the robot has access to its precise grid position and navigates between adjacent cells using predefined motion primitives. This abstraction removes low-level control and estimation noise, allowing us to focus on high-level coordination strategies.

The environment is modeled as an MDP $(\Scal, \Acal, T, R_\mathrm{env}, \gamma)$, where:
\begin{itemize}
    \item $\Scal$ encodes the robot’s position and remaining goals,
    \item $\Acal$ includes movement and image transmission actions,
    \item $T: \Scal \times \Acal \to \Scal$ is a deterministic transition function,
    \item $R_\mathrm{env}: \Scal \to \mathbb{R}$ is the reward function,
    \item $\gamma \in [0,1)$ is the discount factor.
\end{itemize}

At each step, the robot observes its local surroundings up to distance $d$, optionally receives a human-specified trajectory $\zeta_t$, and selects an action $a_t$ to move or transmit an image. 
The human refines their initially inaccurate perception state $x_t \in \Xcal$ over time based on the robot’s actions and communications to provide more accurate guidance.

\begin{problem}\label{problem:conav}
    At each turn, the robot plans over a horizon $H$ by solving:
\begin{subequations}
    \begin{align}
        \max_{a_{0:H-1}} \;
        & \mathop{\mathbb{E}}_{T} \left[ \sum_{t=0}^{H-1} \gamma^t \left(\underbrace{R_{\mathrm{task}}(\tau_{t+1}, \zeta)}_{\text{(1) Task reward}} + \underbrace{\|x_{t+1}-x_t\|}_{\text{(2) Info reward}}\right)\right] \label{obj}\\ 
        \subjectto \quad
        &x_{t+1} \text{ is updated based on } x_t, (\tau_t, a_t), \quad\quad a_t \in \Ocal \label{const:perception_update}\\
        &\tau_{t+1} = \tau_t \oplus T(s_t, a_t), \quad\quad\quad\quad\quad\quad\quad\;\;\; a_t\in \Ucal\label{const:history_update}
    \end{align}
\end{subequations}

The objective~\eqref{obj} maximizes expected cumulative reward under uncertainty in transitions \(T\), combining: 
(1) a \emph{task reward} promoting efficient navigation (\Cref{appendix:task_reward}), and 
(2) an \emph{information reward} measuring perception change via the distance between consecutive perception states.
Constraints~\eqref{const:perception_update}--\eqref{const:history_update} model how robot actions update the human's perception state \(x_t\) and expand the trajectory history \(\tau_t\), with \(\oplus\) denoting sequence concatenation.
\end{problem}

\section{Neural Human Perception Model (NHPM)}
\label{sec:perception_mdp}

\begin{figure*}[ht]
  \centering
  \includegraphics[width=0.9\linewidth,trim=0 70 0 85,clip]{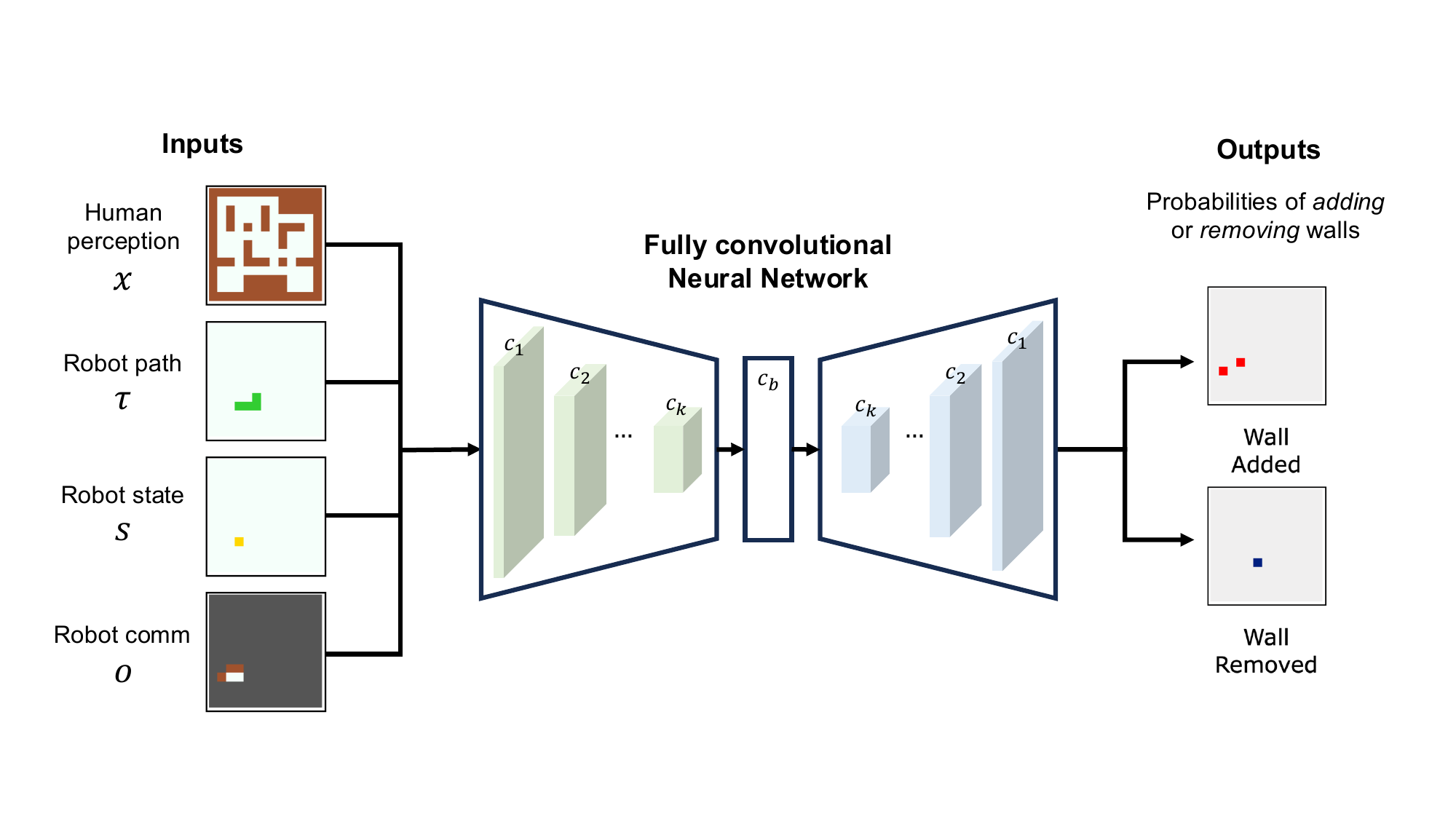}
  \caption{Neural Human Perception Model. \textbf{Inputs:} The human's current perception, the robot's path since the last transmission, and the captured environment grids are processed into 2D masks. \textbf{Outputs:} Probability masks for adding and removing walls.}
  \Description{Diagram of the Neural Human Perception Model with input masks and CNN outputs.}
  \label{fig:nhpm}
  \vspace{-1em}
\end{figure*}

To enable effective human-robot cooperative navigation, we require a model that captures how the human's perceptual state $x_{t+1}$ evolves based on their current state $x_t$ and the robot's communication. As the robot traverses the maze and transmits images from its front camera, the human can update their map of the environment. They may infer that previously assumed obstacles are traversable based on the robot's movements or detect discrepancies between the map and reality from visual observations. We propose to learn this perception model, denoted $F$, from interaction trajectories via
\[
x_{t+1} = F(x_t,\tau_t,o_t),
\]
where \(\tau_t\) is the robot’s recent trajectory and \(o_t\) the latest sent image. 





We propose Neural Human Perception Model (NHPM), a fully convolutional neural network that predicts the human perception transition probabilities modeled in \Cref{sec:perception_mdp}. We denote the model as $F_\theta$ where $\theta$ represents the trainable weights. Such design echoes recent studies of model-based reinforcement learning~\cite{hansen2022temporal}, where the agent first learns the environment dynamics, potentially from image observations~\cite{hafner2019learning,watter2015embed}.

As illustrated in \Cref{fig:nhpm}, our model takes as input the human’s current perception, the robot’s path, and the image captured by the robot, all of which are transformed into a unified 2D representation. These inputs are concatenated along the channel dimension and fed into the CNN, which outputs a two-channel image: one predicting the probability of human adding a new wall and the other predicting the probability of removing a wall. In practice, the model outputs are masked by the current human perception $x$, as a wall cannot be added to a cell that is already labeled as a wall.


We train NHPM from real human data. The crowdsourced dataset consists of 113 episodes from a mapping task, where the robot traverses the maze and stops at random locations and sends a front-view image. The human participants are tasked to update their map to match the real world as closely as possible. 
Detailed information about the dataset is provided in \Cref{sec:dataset_crowdsourcing}.
To enhance generalization, we apply random rotations and flips to the 2D inputs during training, leveraging the orientation invariance of the maze structure.
As will be shown in \Cref{subsec:continuous}, this same crowdsourced dataset can be utilized to train a higher-resolution model capable of planning in continuous environments. Further implementation details are provided in \Cref{appendix:nhpm}.

\section{Information Gain Monte Carlo Tree Search (IG-MCTS)}
\label{sec:method}


To fully solve \Cref{problem:conav}, we propose \emph{Information Gain Monte Carlo Tree Search} (IG-MCTS), a planning framework that jointly optimizes task progress and informative communication. IG-MCTS builds on the NHPM and consists of two main components:
\begin{enumerate}
    \item An augmented reward function that balances task and information objectives. 
    \item An uncertainty-aware MCTS that accounts for unobserved maze regions and stochastic human perception.
\end{enumerate}
    

\subsection{Reward Augmentation}
At each step, IG-MCTS optimizes an augmented reward: $R_\mathrm{aug} = R_{\mathrm{task}} + R_{\mathrm{info}}$. The task reward $R_\mathrm{task}$ drives goal completion and can optionally include heuristics to promote certain robot behaviors. Details of our $R_{\mathrm{task}}$ definition are included in \cref{appendix:task_reward}. The information reward $R_{\mathrm{info}} = \|x_{t+1} - x_t\|$ measures the human’s perceptual update, promoting informative messages. As the robot cannot observe $x$ directly, it estimates $R_{\mathrm{info}}$ using the NHPM (\cref{subsec:uncertainty_aware_simulation}). We use the $\ell_1$ distance for simplicity, as it directly corresponds to the number of cells changed on the human map. Under the uncertainty of the human model, its expected value simplifies to the sum of the predicted probabilities of map edits across all cells.

\subsection{Uncertainty-Aware MCTS}
\label{subsec:uncertainty_aware_simulation}

IG-MCTS follows the standard MCTS steps—\emph{selection}, \emph{expansion}, \emph{rollout}, and \emph{backpropagation}—but extends them to account for uncertainty in both environment dynamics and human perception. In particular, we introduce stochastic simulations during \emph{expansion} and \emph{rollout}, and apply a modified \emph{backpropagation} rule that incorporates transition feasibility.

As detailed in \Cref{algo:IG_MCTS}, both the \emph{expansion} and \emph{rollout} phases involve forward simulation of robot actions. Each tree node $v$ encodes the state $(\tau, x)$, where $\tau$ is the robot’s state history since the last communication and $x$ represents the current human perceptual state. We handle the two action types—movement and communication—differently, as follows:

\textbf{(1)} Movement actions $u$ follow the environment dynamics $T$ defined in \Cref{sec:problem}. The maze layout is observable up to a distance $d$ from previously visited cells, while unexplored areas are treated probabilistically, with a $50\%$ chance of containing walls. During \emph{expansion}, the search node $v'$ from such an uncertain transition is assigned a feasibility score $\delta = 0.5$. During \emph{rollout}, this uncertainty is resolved via sampling: the transition may fail if the target cell is a wall, in which case the robot remains in its current location.

\textbf{(2)} Communication actions $o$ follow the stochastic human perception model $F_\theta$, as introduced in \Cref{sec:perception_mdp}. Since the model defines known transition probabilities, we directly compute the expected information reward $\bar{R}_\mathrm{info}$ as:
\begin{align*}
    \bar{R}_\mathrm{info}(\tau_t, x_t, o_t) &= \mathbb{E}_{x_{t+1}}\left[\|x_{t+1} - x_t\|_1\right] \\
    &= \|p_\mathrm{add}\|_1 + \|p_\mathrm{remove}\|_1,
\end{align*}
where $(p_\mathrm{add}, p_\mathrm{remove}) \gets F_\theta(\tau_t, x_t, o_t)$ denotes the predicted probabilities of adding or removing walls from the human's map. Computing this expectation analytically allows us to estimate node values without relying on repeated visits, improving sample efficiency.




During \textit{backpropagation}, each node's total value $Q(v)$ and visit count $N(v)$ are updated by propagating the sampled return.
For transitions with feasibility $\delta'$ and child return \( q'_{\mathrm{sample}} \), the return used to update the parent node is:
\begin{equation}
    q_{\mathrm{sample}} = r + \gamma \left[ \delta' \, q'_{\mathrm{sample}} + (1 - \delta') \frac{Q(v)}{N(v)} \right],
\end{equation}
where $\frac{Q(v)}{N(v)}$ serves as a fallback estimate for the current value at the parent node.

\section{Experiments}\label{sec:exp}
This section starts with an evaluation of NHPM’s accuracy in predicting human map updates (Section~\ref{subsec:nhpm_eval}), then presents a within-subject study comparing IG-MCTS to teleoperation and instruction-following on cognitive load and task performance (Section~\ref{subsec:user_study}).

\subsection{Human Perception Model Evaluation}\label{subsec:nhpm_eval}
We evaluate the effectiveness of the proposed NHPM in predicting how humans perceive environmental information based on a robot’s movement and communication.

\begin{figure*}
    \centering
    \includegraphics[width=\linewidth]{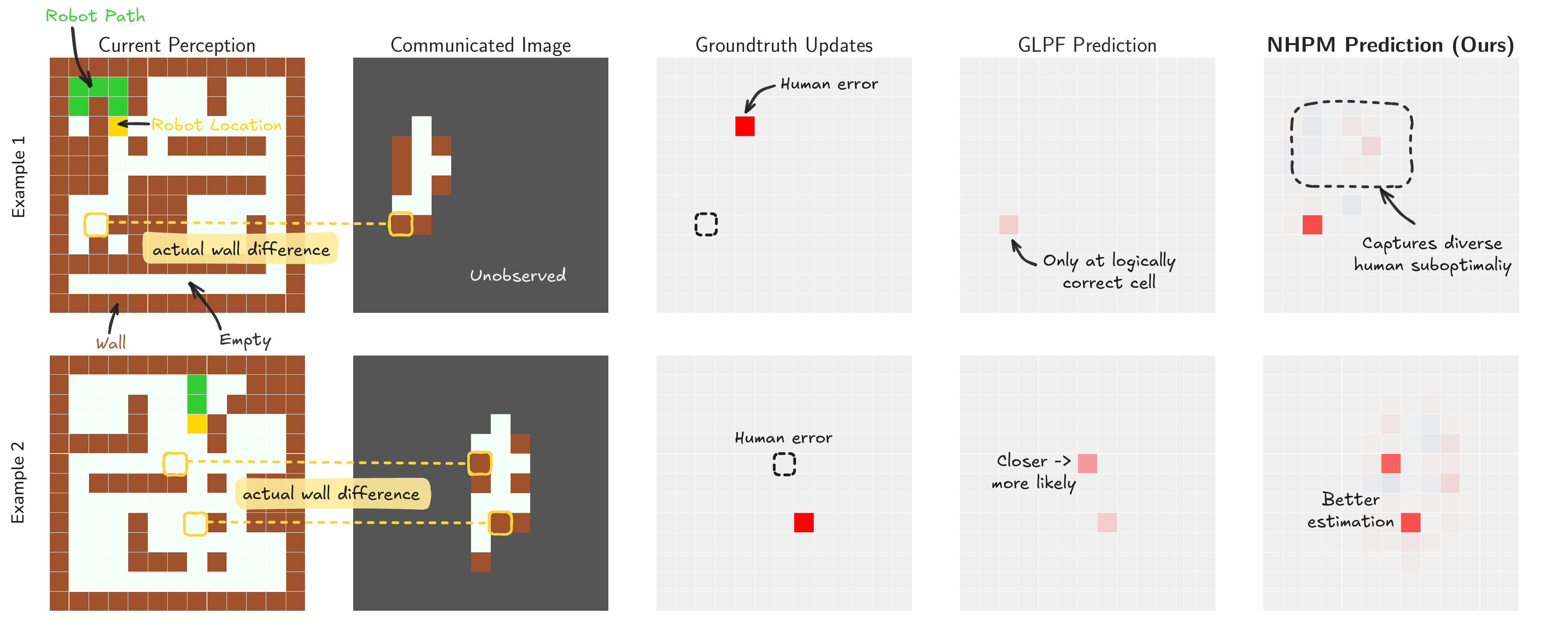}
    \begin{minipage}{0.38\linewidth}
        \centering
        \vspace{-10pt}
        \includegraphics[width=\linewidth]{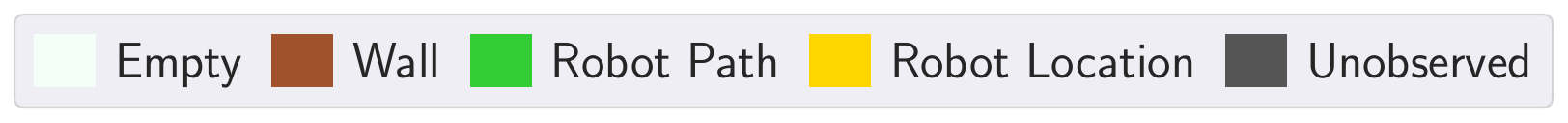}
    \end{minipage}
    \hspace{0.02\linewidth}
    \begin{minipage}{0.52\linewidth}
        \centering
        \vspace{3pt}
        \includegraphics[width=\linewidth]{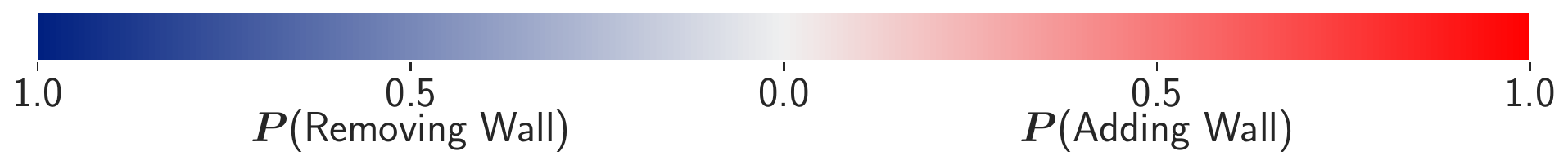}
    \end{minipage}
    \hspace{3mm}
    \vspace{-2mm}
    \caption{Visualization of human perception models. The left two columns show inputs: the human's current map, the robot's path, and the visible grids communicated by the robot. The models predict how the human will update the maze based on this information. \textbf{Top:} The human mistakenly marks a nearby wall at the wrong location. Despite never encountering this exact scenario, NHPM successfully anticipates the error by generalizing from similar training examples. \textbf{Bottom:} The human correctly adds a distant wall instead of a nearby one, a behavior accurately predicted by NHPM.}
    \Description{Visualization comparing predicted and actual human map updates in maze navigation tasks.}
    \label{fig:GLPF_vs_NHPM}
    \vspace{-1em}
\end{figure*}

\paragraph{Baseline: Grid-Based Logistic Psychometric Function.}
We compare our method to the Logistic Psychometric Function (LPF), a standard model used to relate human observer performance (e.g., detection or discrimination) to stimulus intensity~\cite{treutwein1999fitting,wichmann2001psychometric,klein2001measuring}. LPF fits a curve that predicts responses to a single, scalar stimulus and is therefore limited in its ability to capture complex spatial dependencies.

In our case, the human receives visual input from images, where stimulus is distributed across many spatial locations. Since LPF cannot directly handle such structured input, we adapt it to operate at the grid-cell level, treating each cell independently. We define the stimulus intensity for a cell at distance $d$ from the robot as $x(d) = e^{-\alpha d}$, assuming that correctly labeled cells provide no stimulus. 

This adapted model, which we call the Grid-Based LPF (GLPF), estimates the probability that the human updates their belief about a grid cell given its stimulus intensity $x$:
\begin{equation}
P(y = 1 \mid x) = \rho + \frac{1 - \rho - \lambda}{1 + e^{-\beta (x - \alpha)}},
\end{equation}
where $\beta$ controls sensitivity to stimulus changes and $\alpha$ sets the midpoint of the logistic curve. The parameter $\rho$ represents the guessing rate—the probability of a response in the absence of stimulus—while $\lambda$ accounts for lapses, the chance of missing at maximum intensity.

\paragraph{Evaluation.}
We split the dataset into training and test sets and consider three distinct test settings:
\begin{enumerate}[leftmargin=*]
    \item GLPF-Train: The psychometric function is fit on the training set to evaluate how well it generalizes to unseen environments based on prior human data.
    \item GLPF-Test: To establish an upper performance bound, we fit the GLPF directly to the test set. This removes the generalization gap, revealing the best-case scenario for an LPF-based approach.
    \item NHPM: We train the neural network using backpropagation, optimizing parameters by minimizing the binary cross-entropy loss between predicted and ground-truth human edits.
\end{enumerate}

\paragraph{Quantitative Results}
The test loss and accuracy in \Cref{tab:perception_dynamics_evaluation} highlight NHPM's advantage over GLPF, demonstrating that incorporating spatial structure and contextual awareness improves human perception prediction. Even when fit on test data, GLPF remains limited by its lack of spatial expressiveness, whereas the CNN generalizes effectively from the training set.
\begin{table}[ht]
\begin{center}
\begin{small}
\begin{tabular}{c@{\hskip 8pt}c@{\hskip 8pt}c@{\hskip 8pt}c@{\hskip 8pt}r}
\toprule
Method & \makecell{Train Loss \\(MBCE)} & \makecell{Test Loss \\(MBCE)} & \makecell{Test Accuracy \\ (IoU @ $\Gamma$)} \\
\midrule
GLPF-Train & $1.15 \cdot 10^{-1}$ & $9.10 \cdot 10^{-2}$  & $0.335 \, @ \, 0.38$ \\
GLPF-Test & N/A & $8.98 \cdot 10^{-2}$ & $0.335 \, @ \, 0.21$ \\
NHPM (Ours)  & $1.36 \cdot 10^{-2}$ & $\mathbf{1.43 \cdot 10^{-2}}$ & $\mathbf{0.352}$  \\
\bottomrule
\end{tabular}
\end{small}
\end{center}
\caption{NHPM achieves lower mean binary cross entropy error (MBCE) and higher prediction accuracy measured in intersection over union (IoU) on the test set than the baseline method, even when GLPF is optimized on the test set with an ideal decision boundary~$\Gamma$. 
}
\label{tab:perception_dynamics_evaluation}
\vspace{-1.5em}
\end{table}

\paragraph{Qualitative Analysis}
In \Cref{fig:GLPF_vs_NHPM}, we visualize the outputs of human perception models and highlight two representative scenarios where NHPM outperforms GLPF. In the top row, the human mistakenly marks a wall close by, misjudging its distance from the first-person view image. Despite never encountering this exact test scenario, NHPM correctly anticipates the error by generalizing from similar patterns in the training set. In the bottom row, the robot captures an image down a hallway, and the human adds a distant wall instead of a nearby one, likely because it appears in the center of the image and aligns with an existing wall on the map. NHPM accurately predicts this behavior, while the psychometric function assigns a low probability due to the wall's distance.

\subsection{User Study}\label{subsec:user_study}

We investigate whether IG-MCTS can reduce human cognitive load while maintaining task performance comparable to teleoperation and instruction-following with a within-subject user study.

\paragraph{Independent Variables.}  
The study compares IG-MCTS to two baseline interaction methods:

\begin{itemize}
    \item \emph{Teleoperation:} Participants manually control the robot’s low-level actions \(u_t^\mathrm{low} \in \Ucal^\mathrm{low}\) at each timestep using keyboard arrows. The robot executes actions deterministically based on the low-level transition function \(T_\mathrm{low}\) and streams real-time RGB images from its front camera.
    \item \emph{Instruction-following:} Participants specify a trajectory $\zeta = \langle s_t, \allowbreak\dots, s_{t+n} \rangle$. The robot autonomously follows this trajectory based on the transition function \(T\). If blocked, the robot pauses and prompts humans for updated guidance. RGB image streaming is provided for teleoperation.
\end{itemize}

\paragraph{Dependent Measures.}  
We measured task performance (robot steps, human guidance instances, mapping accuracy) and physiological cognitive load via eye‐tracking (pupil diameter measurements, blink rate, and fixation shifts between areas of interest).

\emph{Why these eye tracking metrics?}
Task-evoked pupillary responses are well-established indicators of mental effort~\cite{beatty1982task,hess1964pupil}, with greater cognitive demand leading to larger pupil dilation.
We measure mean pupil diameter and percent change in pupil dilation (PCPD)~\cite{kruger2013measuring}.
Blink rate, on the other hand, is inversely correlated with cognitive load, with higher rates indicating reduced mental effort~\cite{zagermann2018studying,zagermann2016measuring}.
Fixation shifts between areas of interest (AOIs) reflect visual and cognitive resource allocation during tasks~\cite{joseph2020potential}. Fewer shifts suggest reduced effort in integrating information across AOIs.

\begin{figure*}
    \centering
    \begin{minipage}{0.45\linewidth}
        \centering
        \includegraphics[width=\linewidth]{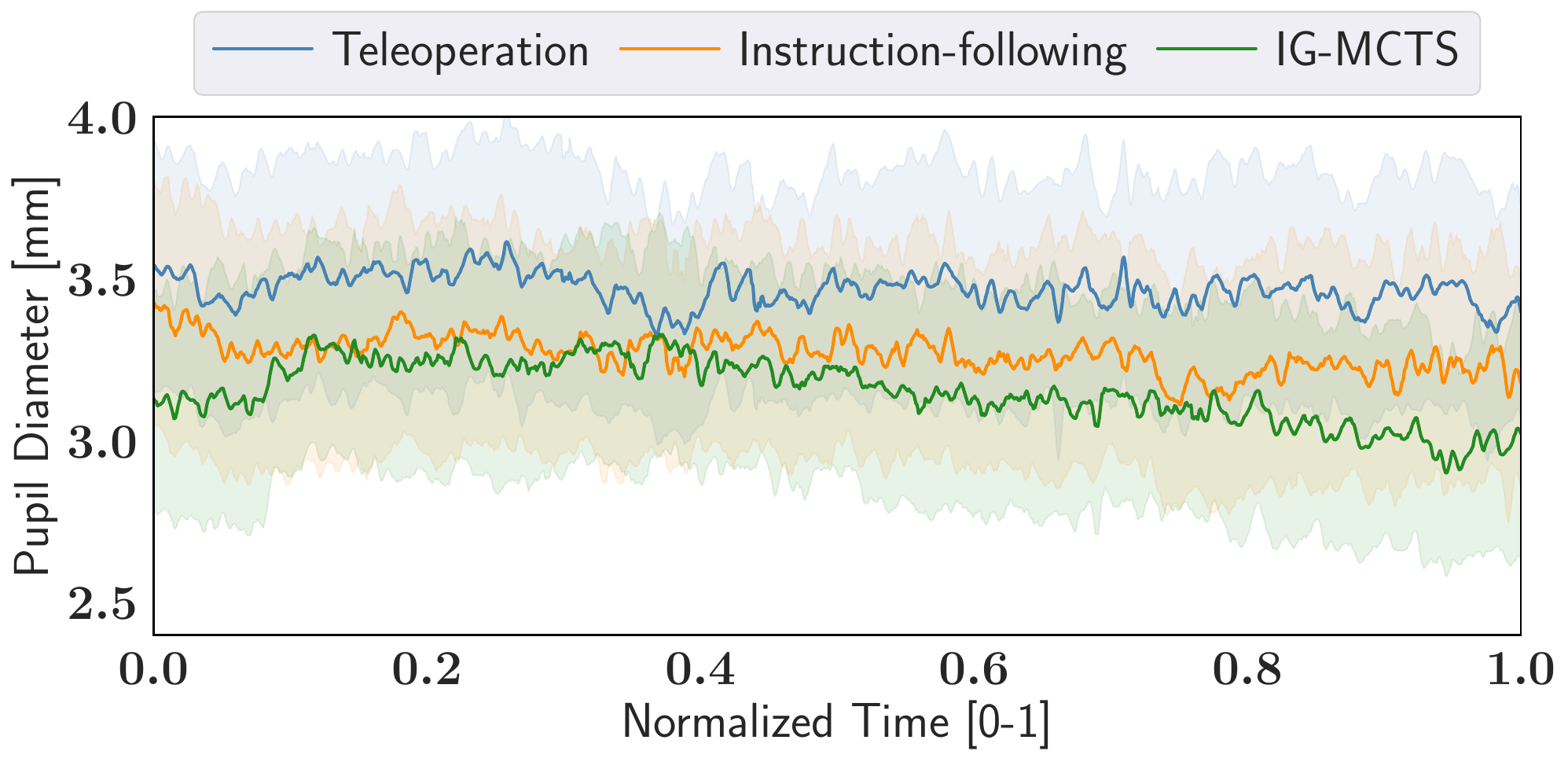}
        \vspace{-2.5em}
        \caption{Mean pupil diameter with $95\%$ CI (Interpolation: $1000$, smoothing: $5$).}
        \label{fig:pupil_diameter_plot}
    \end{minipage}
    \hspace{2em}
    \begin{minipage}{0.45\linewidth}
        \centering
        \begin{small}
        \resizebox{\linewidth}{!}{
        \begin{tabular}{l@{\hskip 3pt}c@{\hskip 3pt}c@{\hskip 3pt}c}
        \toprule
        Method & \makecell{PCPD $\downarrow$} & \makecell{Blink Rate $\uparrow$} & \makecell{Fixation Shift Rate $\downarrow$} \\
        \midrule
        Teleoperation & $31.11\pm7.66$ & $8.84\pm6.15$  & $41.00\pm12.27$ \\
        \makecell{Instruction-\\following} & $20.27\pm6.80$ & $10.20\pm5.49$ & $33.91\pm7.07$ \\
        IG-MCTS  & $\mathbf{16.80\pm7.33}$ & $\mathbf{12.42\pm8.54}$ & $\mathbf{32.73\pm7.21}$ \\
        \bottomrule
        \end{tabular}
        }
        \end{small}
        \caption{Eye-tracking metrics of cognitive load: PCPD (\%), blink rate (/min), and fixation shift rate (/min).}
        \label{tab:eye_metrics}
    \end{minipage}
\end{figure*}
\begin{figure*}[h]
\vspace{-1.5em}
\begin{center}
    \includegraphics[width=\linewidth]{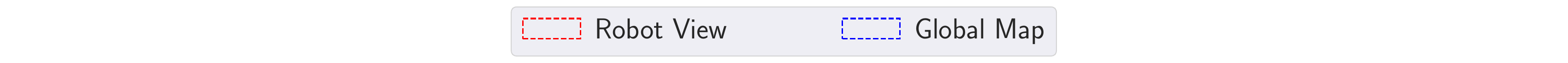}
    \begin{minipage}[b]{0.3\textwidth} 
        \centering
        \includegraphics[width=\linewidth]{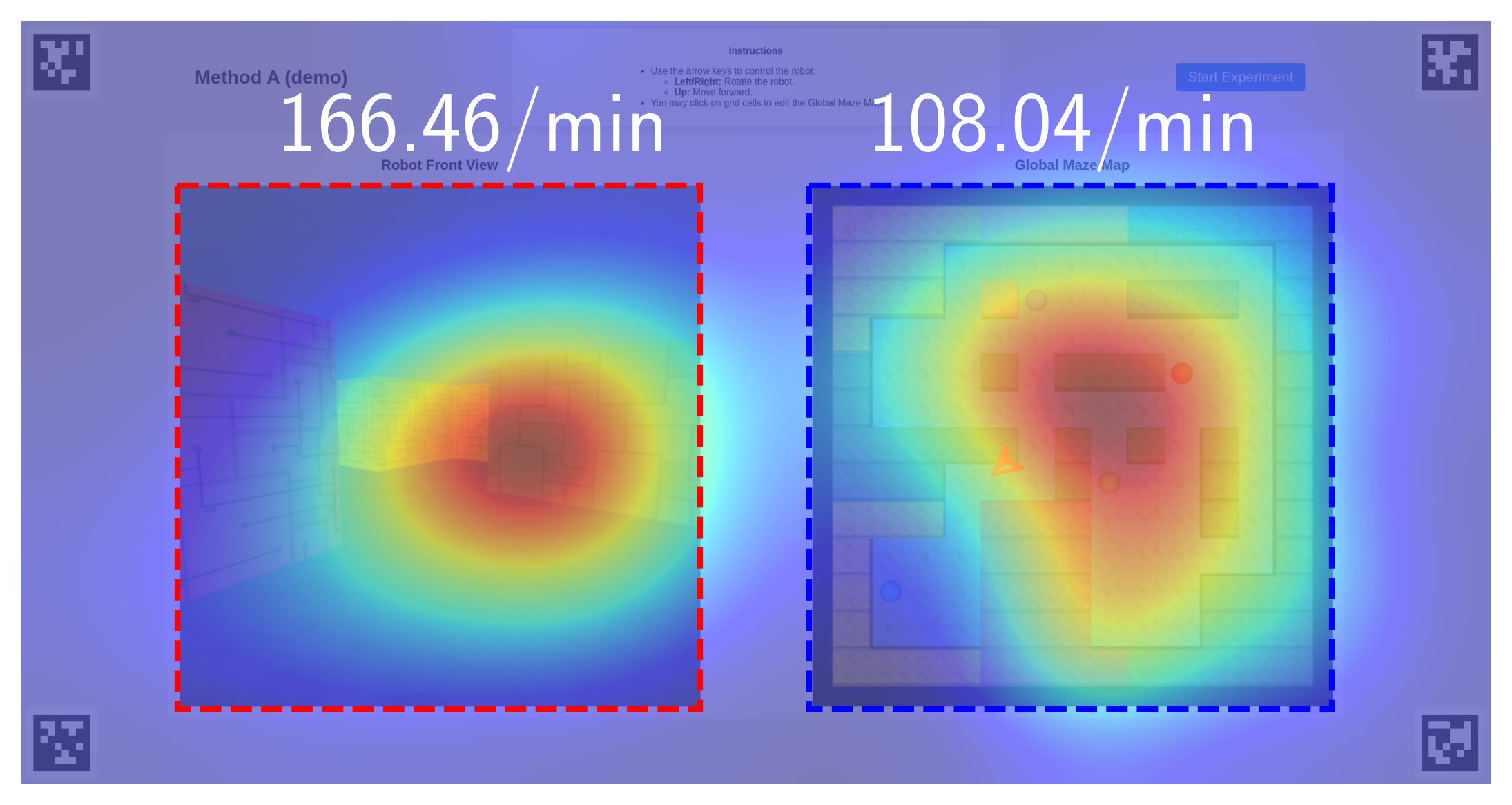}
        \caption*{(a) Teleoperation}
    \end{minipage}
\hfill
    \begin{minipage}[b]{0.3\textwidth}
        \centering
        \includegraphics[width=\linewidth]{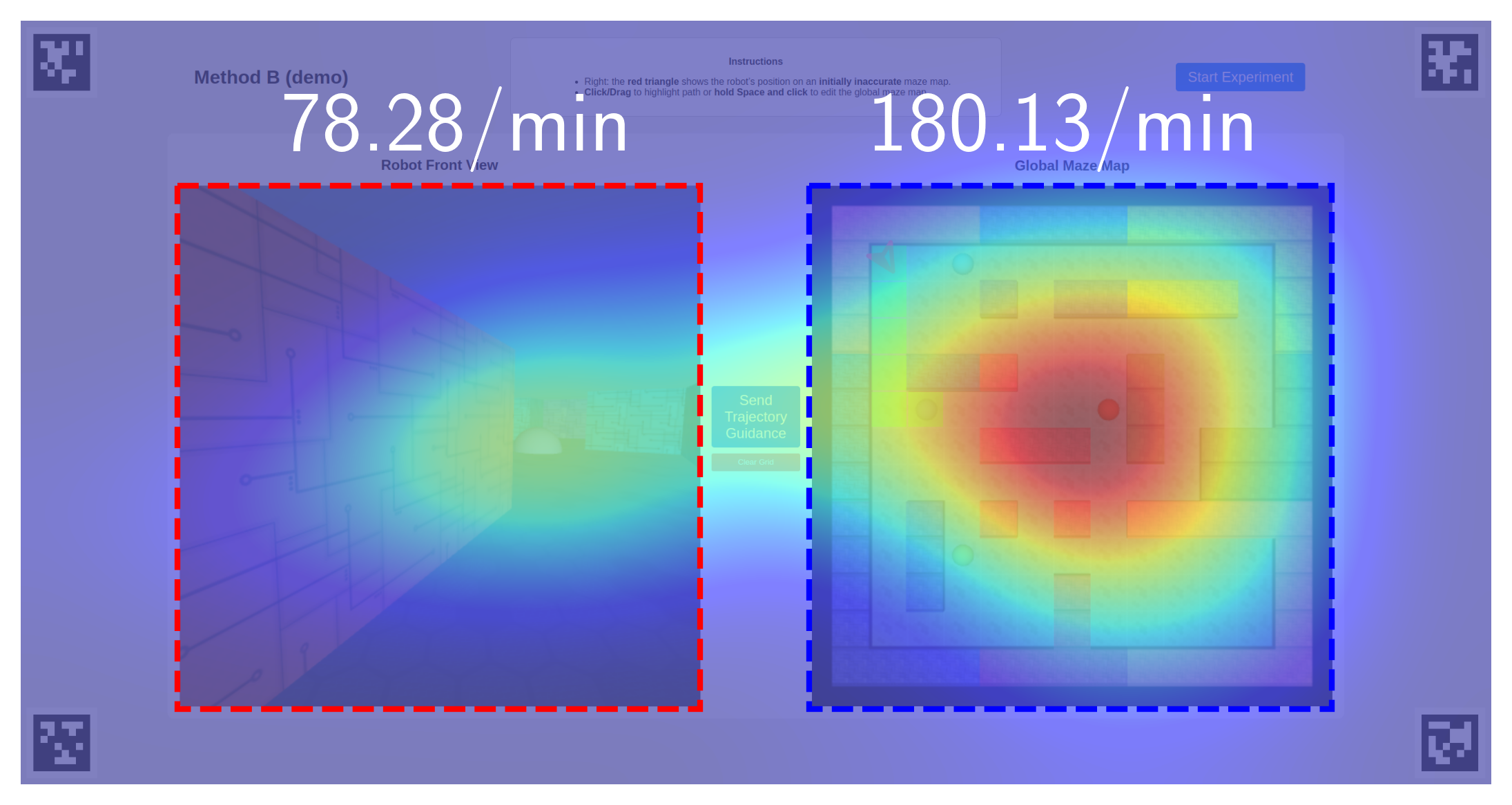}
        \caption*{(b) Instruction-following}
    \end{minipage}
    \hfill
    \begin{minipage}[b]{0.3\textwidth}
        \centering
        \includegraphics[width=\linewidth]{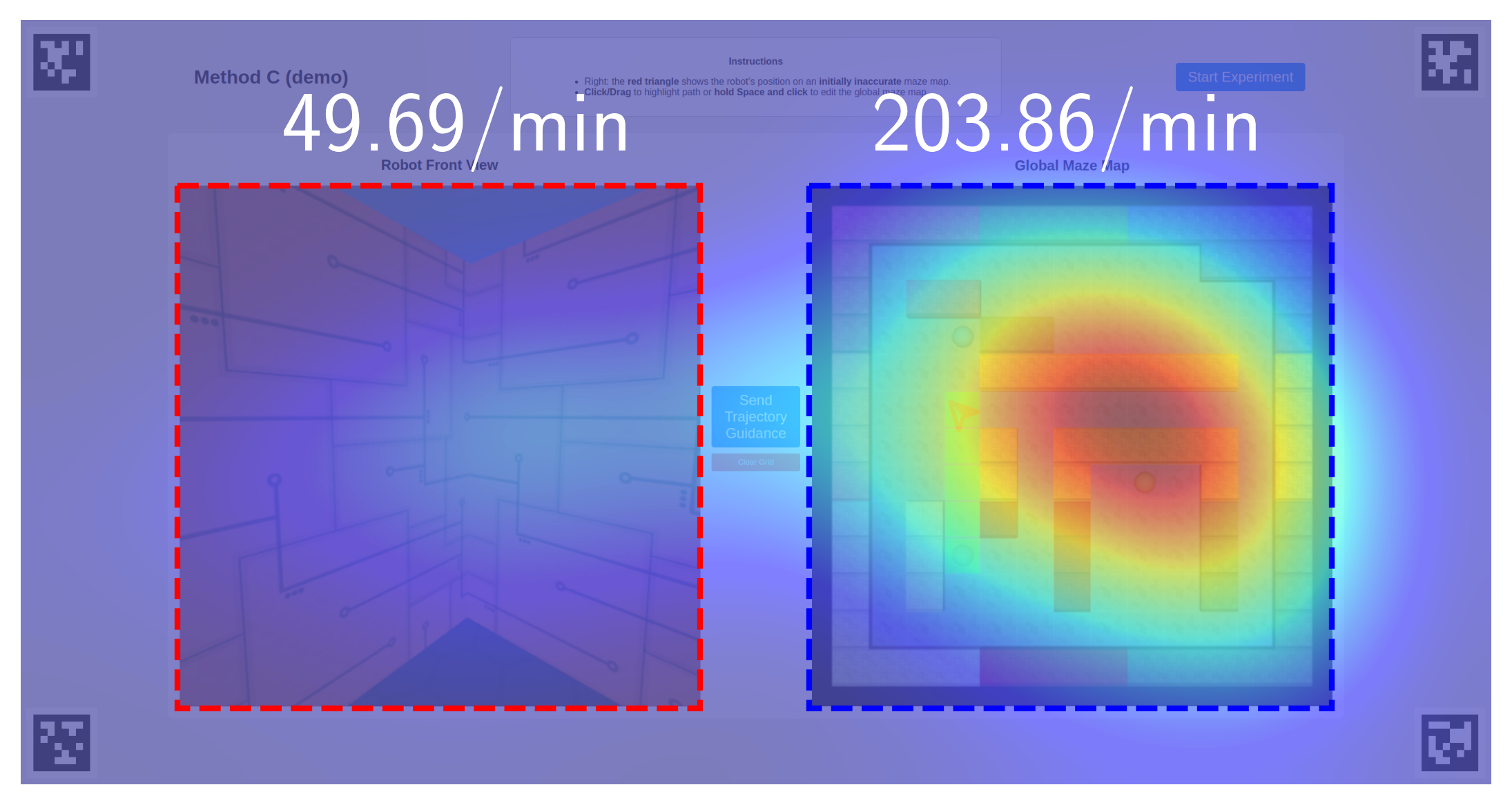}
        \caption*{(c) IG-MCTS}
    \end{minipage}
\caption{Aggregate gaze heatmaps showing visual attention distribution. AOIs: the robot ego-view (dashed red box) and the top-down global map (dashed blue box). Fixation rates in white text.}
\label{fig:gaze_heatmaps}
\end{center}
\vspace{-1.5em}
\end{figure*}

\paragraph{Hypotheses.} 
We hypothesize that IG-MCTS will (\textbf{H1}) yield eye-tracking metrics indicative of lower cognitive load than teleoperation, and (\textbf{H2}) require significantly less communication while maintaining comparable task performance to instruction-following.

\paragraph{Participants.}
The study recruited $14$ graduate students, with a demographic breakdown of $78.57\%$ male, $21.43\%$ female. The participants' average age was $26.79$ years (SD = $3.17$). 

\paragraph{Procedure.}
Participants wear a Pupil Labs Core eye tracker~\cite{kassner2014pupil} and complete a 5-point calibration before the study. Each participant completes three randomized sessions, corresponding to different interaction types, to control for ordering effects~\cite{martin2007doing}. The experiment is conducted under controlled lighting with emotionally neutral content to minimize confounding factors in pupilometry.
Each session begins with a baseline pupil diameter measurement (up to 30 seconds) with participants reading a brief, neutral paragraph (\Cref{appendix:baseline_pupil}). Participants then practice the session-specific controls in a demo maze before completing tasks in three distinct maze layouts. The same layouts are reused across sessions, with $90^\circ$ or $180^\circ$ rotations to prevent memorization and maintain engagement.



\subsubsection{Results}

\paragraph{On H1 (Eye tracking Metrics for Cognitive Load).}
To evaluate hypothesis \textbf{H1}, we examine eye-tracking metrics indicative of cognitive load: pupil dilation, blink rate, and fixation jump rate. 
A repeated-measures ANOVA\footnote{We report F-statistic with degrees of freedom $F(df)$, significance level $p$ and effect size $\eta^2$.} is conducted to assess overall differences across the three methods, followed by paired $t$-tests\footnote{We report $t$-statistic with degrees of freedom $t(df)$, significance level $p$, and Cohen’s $d$ as effect size.} for post-hoc comparisons.

\Cref{fig:pupil_diameter_plot} presents mean pupil diameter curves for each method, normalized to a $0$-time scale, interpolated over $1000$ points, and smoothed with a window size of $5$. IG-MCTS consistently exhibits the smallest pupil diameter, followed by instruction-following, while teleoperation shows the largest diameter overall.
This pattern aligns quantitatively with the percent change in pupil diameter (PCPD) statistics reported in \Cref{tab:eye_metrics} (calculation details in \Cref{appendix:pcpd_calculation}). 
A repeated-measures ANOVA reveals significant differences on pupil dilation among the three methods ($F(2,26)=6.045, p=0.005, \eta^2=0.317$). Post-hoc comparisons further show that IG-MCTS yields significantly lower pupil dilation than teleoperation ($t(13) = 2.807$, $p = 0.015$, $d = 0.750$), suggesting that participants experienced reduces cognitive effort when collaborating with IG-MCTS.

\Cref{tab:eye_metrics} reports the blink rate across methods. Although IG-MCTS yields a numerically higher blink rate compared to the two baselines, a repeated-measures ANOVA finds no overall significant difference among the three methods ($F(2,26) = 0.904$, $p = 0.413$, $\eta^2 = 0.065$). 
Nevertheless, a pairwise comparison reveals that IG-MCTS results in a significantly higher blink rate than teleoperation ($t(13) = -3.658$, $p = 0.003$, $d = -0.978$). This result indicates the nuanced nature of interpreting blink rate as an indicator of cognitive load, as suggested in prior work \cite{magliacano2020eye,ledger2013effect}.

Further, \Cref{fig:gaze_heatmaps} shows the aggregate gaze heatmaps illustrating visual attention distribution between robot ego-view and global map across methods. Teleoperation divides participants' attention between both views due to the continuous demands of manual control. Instruction-following reduces these demands, shifting greater attention toward the global map. IG-MCTS further focuses participants’ attention almost exclusively on the global map by automating low-level control. 
This trend is quantitatively supported by the fixation rates for each area of interest (AOI), labeled in white text within \Cref{fig:gaze_heatmaps}, and by the fixation shift rates reported in \Cref{tab:eye_metrics}. A repeated-measures ANOVA shows no overall significant difference in fixation shift rate across methods ($F(2,26) = 3.091$, $p = 0.057$, $\eta^2 = 0.192$). However, pairwise comparisons reveal that IG-MCTS results in significantly fewer fixation shifts compared to teleoperation ($t(13) = 2.676$, $p = 0.019$, $d = 0.715$).

Overall, these results provide support for \textbf{H1}, indicating that IG-MCTS tends to reduce cognitive load compared to teleoperation.


\begin{table}[ht]
\begin{center}
\resizebox{\linewidth}{!}{
    \begin{tabular}{lcc}
    \toprule
    Method & Communication (MB) & \#Robot Step \\
    \midrule
    Teleoperation & $107.18\pm 38.50$ & $609.71\pm219.01$ \\
    Instruction-following & $84.81\pm \phantom{0}6.15$ & $482.45\pm\phantom{0}34.99$ \\
    IG-MCTS  & $\mathbf{\phantom{0}2.38\pm \phantom{0}0.90}$ & $\mathbf{457.36\pm118.48}$ \\
    \bottomrule
    \end{tabular}
}
\end{center}
\caption{Task metrics (mean~$\pm$~standard deviation). Communication assumes $180$ KB/image.}
\label{tab:task_metrics}
\vspace{-2em}
\end{table}

\paragraph{On H2 (Task Metrics).} 
\Cref{tab:task_metrics} summarizes task performance metrics averaged across all participants and both maze layouts. 
As hypothesized (\textbf{H2}), IG-MCTS drastically reduces communication overhead---transmitting only $2.38\pm0.90$ MB on average---compared to continuous teleoperation ($107.18\pm38.50$ MB) and instruction‐following ($84.81\pm6.15$ MB).
At the same time, IG‐MCTS takes slightly fewer robot steps ($457.36\pm118.48$) than instruction‐following ($482.45\pm34.99$) and substantially fewer steps than teleoperation ($609.71\pm219.01$).
Paired $t$-tests confirm that the reduction in steps is highly significant both against instruction-following ($t(13)=46.71$, $p<0.001$, $d=12.96$) and teleoperation ($t(13)=9.85$, $p<0.001$, $d=2.73$). 
The results suggest that IG-MCTS achieves fewer or comparable robot steps to instruction-following with significantly reduced communication, providing support for \textbf{H2}.

\paragraph{Key Takeaways.}
We observe clear quantitative trends: IG-MCTS achieves significant improvements in eye-tracking metrics---pupil responses and fixation shift rates---over teleoperation, likely due to its selective transmission of task-relevant visual information rather than continuous streaming. It also reduces communication overhead by over $97\%$ relative to both baselines while maintaining comparable robot steps.
Qualitatively, we observe several notable IG-MCTS behaviors during interaction: it proactively moves toward visible goals based on accumulated SLAM observations, even without explicit human guidance; it strategically pauses and reorients to gather critical information when human guidance is suboptimal; and it improves communication efficiency by angling itself to capture broader scene information within a single snapshot.


\subsection{Illustrative Continuous-Space Example}\label{subsec:continuous}
\begin{figure*}
    \centering
    \includegraphics[width=0.85\linewidth,trim=30 50 30 50,clip]{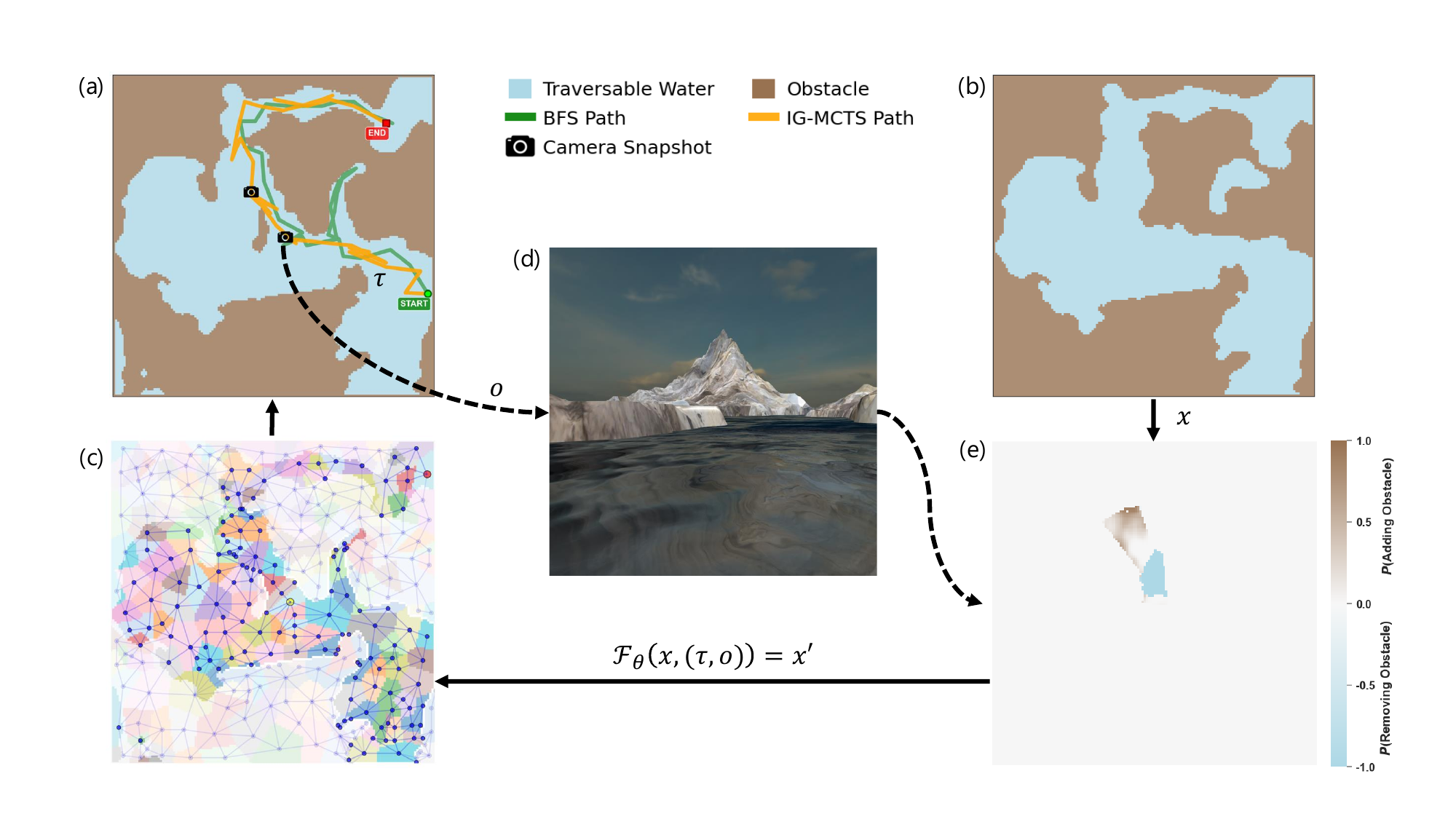}
    \caption{
    (a) Ground-truth terrain map showing the greedy BFS trajectory (green) and the IG-MCTS trajectory (orange). 
    (b) The initial map provided to the human operator and also shared with the robot, which differs from the ground-truth map in (a). 
    (d) The robot stops and chooses  a direction to share its ego-view image that maximizes information gain for the human. 
    (e) NHPM predicts how the human updates their map after observing the shared photo shown in (d). 
    (c) The Voronoi-partitioned graph on which IG-MCTS performs planning. The current agent location is marked in yellow and the goal is marked in red. The graph is constructed dynamically as the agent observes the environment. Similar to the discrete setting, unobserved regions are assumed traversable with a cost. The graph is accurately separated by observed obstacles.
    }
    \label{fig:continuous}
    \vspace{-1em}
\end{figure*}

Until now, we have focused on discrete, grid-based environments. However, many real-world navigation tasks occur in continuous domains, where motion and perception are not confined to discrete cells.
To demonstrate the broader applicability of the proposed framework, we present an illustrative continuous-space example shown in Figure \ref{fig:continuous}. This example is derived from a synthetic terrain generated using a smooth height map, where low-elevation areas are filled with water and treated as traversable by boat; high elevations represent obstacles.
A start and a goal are placed at opposite ends to require long-range planning across complex terrain that captures essential aspects of real-world continuous navigation while remaining simple enough to visualize and analyze.

\paragraph{NHPM in Continuous Space.}
While Section~\ref{sec:perception_mdp} introduced the Neural Human Perception Model (NHPM) in the context of a $13\times13$ discrete maze, the same formulation naturally extends to continuous navigation settings. The core idea—learning a mapping $F_\theta(x_t,\tau_t,o_t)$ that predicts how a human’s internal map evolves given their current belief, the robot’s trajectory, and recent visual input—remains unchanged. 
In continuous settings, the environment is represented as a smooth terrain height map rasterized to $150\times150$ pixels. As a result, we maintain a $150\times150$ binary mask denoting which pixelated regions are traversable.
NHPM is implemented as a convolutional encoder–decoder operating on this larger field. We reuse the crowd-sourced mapping data to train this model. Because of the increased size and representation flexibility, we apply extensive data augmentation to provide richer training examples: we scale up the $13\times13$ maps to random sizes between $100$ and $150$ with random rotation (0--360$^\circ$) and axis flips before embedding each sample on a $150\times150$ canvas to promote orientation invariance.
Training uses binary cross-entropy with class weighting to counteract the sparsity of human updates. We observe that participants edit around 1 cell on average per robot communication, resulting in the majority of training labels being negative (no edit). Class weighting effectively emphasizes positive labels, and prevents the model from collapsing into predicting no edits all the time. Full network and training details are in \Cref{appendix:nhpm}.

\paragraph{IG-MCTS on Voronoi Graphs.}
To facilitate planning in a continuous terrain, the environment is discretized into a connectivity-preserving graph using Voronoi partitioning \cite{aurenhammer1991voronoi} constrained by the agent's current knowledge of the environment. The unobserved areas are assumed traversable with a 50\% probability. A set of seeds is generated by sampling points within the navigable space, with a controlled fraction placed near obstacle boundaries to capture narrow passages and the remainder distributed across open areas. Each traversable cell is assigned to its nearest seed, forming discrete Voronoi regions that reflect local spatial influence. Adjacent regions are then connected through 4-neighbor boundary analysis, where edges are added only if the connecting line segment remains entirely within the traversable area. This results in an undirected graph whose nodes represent region centroids and edges encode valid motion primitives between neighboring regions. The Information Gain Monte Carlo Tree Search (IG-MCTS) algorithm operates on this graph structure, expanding nodes based on both the expected information gain (computed from the NHPM belief model) and the feasibility of traversing between connected regions. When a movement action is chosen, the agent hops to the chosen neighboring cell and observes its surroundings up to a distance of 50 pixels. The observation circle is also naturally blocked by obstacles to ensure realism. If the observations result in an update to the agent's knowledge of the environment, the traversable graph is re-constructed to reflect the up to date knowledge.


As shown in \Cref{fig:continuous}, the ground-truth terrain map in (a) differs from the initial human-perceived map in (b). The robot is aware of the human's initial map, and can estimate the potential information gain for the human if a photo of the environment is shared.
Without human input and acting greedily via breadth-first search (BFS) on the traversable graph, the robot’s trajectory (green) tends to follow locally optimal paths that eventually lead to a dead-end.
With IG-MCTS, however, the robot continuously follows human guidance, estimates the potential information gain from a communication action and balances the goal-reaching reward.
When the information gain outweighs a fixed communication cost, the robot pauses at its current position, chooses the most informative angle, and shares its current ego-view, as illustrated in (d).
Based on a given viewpoint, the trained NHPM predicts how the human is likely to update the map, as shown in (e). More pixels flipped in expectation are considered higher information gain. After a communication step, the robot receives a new trajectory guidance from the human and re-plans its motion over the Voronoi-partitioned graph in (c), enabling it to continue along the more efficient trajectory (orange) in (a). Overall, IG-MCTS not only produces more efficient paths but also highlights critical map discrepancies for the human operator, enabling more effective collaborative navigation workflow.
\section{Conclusion}\label{sec:conclusion}

We address the challenge of human–robot cooperative navigation under incomplete information by introducing Information Gain Monte Carlo Tree Search (IG-MCTS). By leveraging a learned Neural Human Perception Model (NHPM), IG-MCTS optimizes information sharing based on predicted human belief updates. User studies show that IG-MCTS substantially reduces communication demands and cognitive load while maintaining task performance comparable to teleoperation and instruction-following baselines. We further demonstrate generalization beyond discrete mazes through a continuous-space example, where NHPM scales via a deeper encoder–decoder and IG-MCTS operates over a dynamically constructed Voronoi-partitioned traversability graph. Future extensions may include improving NHPM’s generalization through meta-learning or domain adaptation, and leveraging its alternating execution–communication cycle to enable scalable multi-robot supervision in domains such as search and rescue~\cite{murphy2004human}.

\newpage
\bibliographystyle{ACM-Reference-Format} 
\bibliography{ref}


\newpage
\appendix
\onecolumn
\section*{\LARGE Appendix}

\section{Dataset Crowdsourcing}
\label{sec:dataset_crowdsourcing}
We collect data through a mapping task designed in the CoNav-Maze environment, where participants update an initially inaccurate map to reflect the true maze layout. Each time a participant presses a button, the robot autonomously navigates to a target location and captures an image from a random viewpoint.
The robot has access to both the ground-truth environment and the participant's current belief map. It first visits a set of predefined goal locations, then continues by selecting random grid cells as new targets, prioritizing areas with discrepancies between the participant's map and the true layout.

\begin{figure}[ht]
    \centering
    \includegraphics[width=\linewidth,trim=0 10 0 10,clip]{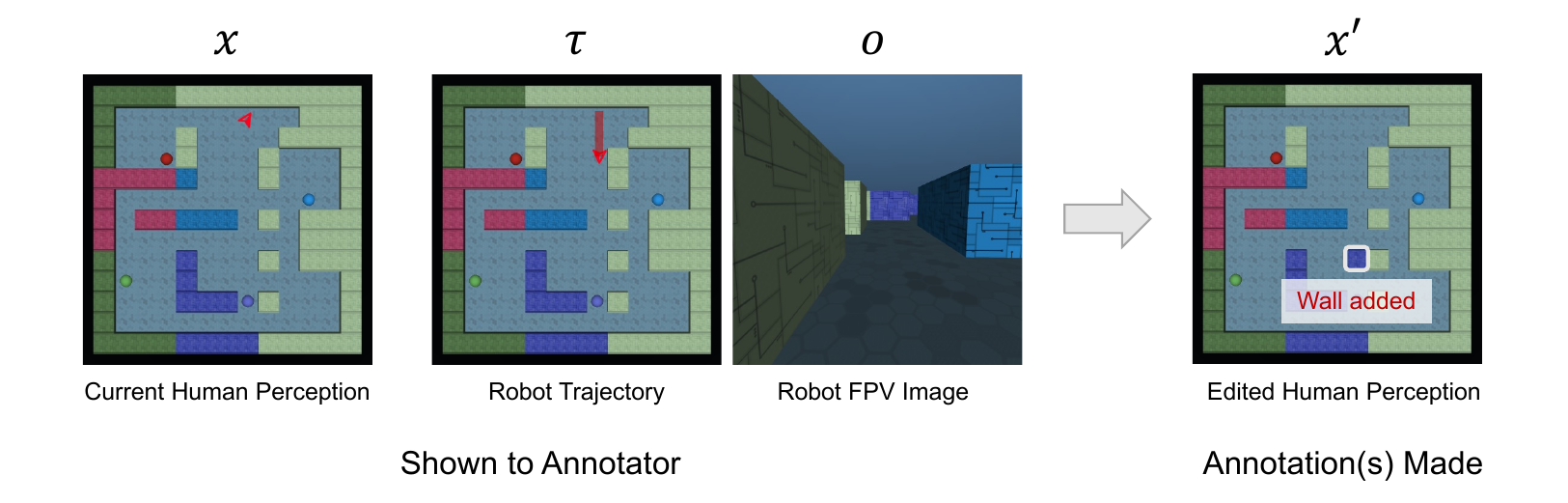}
    \caption{An example of a training segment used to model human perception. Each segment corresponds to a robot communication event and is represented as a tuple $(x, \tau, o, x')$, where $x$ is the current human belief map, $\tau$ is the robot's trajectory since the last communication, and $o$ is the first-person view (FPV) image transmitted to the human. The human then updates their map, producing a new belief state $x'$ (e.g., adding a wall).}
    \label{fig:enter-label}
\end{figure}


We recruit over $50$ annotators via the Prolific platform~\cite{palan2018prolific}. Each participant completes the mapping task on three randomly generated mazes. Participants may proceed to the next maze once the robot reaches all four goal locations but can optionally continue refining their maps. To encourage accurate annotations, participants receive a performance-based bonus proportional to their final mapping accuracy.

The resulting dataset comprises $113$ valid trajectories with final mapping accuracy above $50\%$, which we use for training and evaluation via a $95/5$ train/test split. Each trajectory corresponds to a distinct randomly generated maze, ensuring diversity in both spatial structure and human annotation behavior. An additional $254$ trajectories are excluded due to poor engagement or trivial performance (accuracy $\leq 50\%$).

Each valid trajectory is segmented at robot communication events to form individual training samples. Each segment is represented as a tuple $(x, \tau, o, x')$, where $x$ is the participant’s belief map, $\tau$ the robot’s motion path since the last communication, and $o$ the image shared with the participant. The resulting dataset contains $1{,}505$ training segments and $137$ test segments.





\newpage
\section{NHPM Implementation}\label{appendix:nhpm}

\paragraph{Model Architecture and Hyperparameters.}
The Neural Human Perception Model (NHPM) is implemented as a fully convolutional network that predicts probability masks for adding and removing walls (or traversable regions). Table~\ref{tab:arch_params} summarizes the main architectural and training parameters for both the discrete and continuous versions.

\begin{table}[h!]
\centering
\caption{Summary of NHPM architectures and training parameters.}
\label{tab:arch_params}
\begin{tabular}{lcc}
\toprule
\textbf{Map Type} & \textbf{Discrete (13$\times$13)} & \textbf{Continuous (rasterized to 150$\times$150)} \\
\midrule
Architecture type & Fully convolutional (FCN) & Encoder--decoder \\
Convolution layers & 4 $\times$ Conv($3\times3$, stride 1) & ($6 + 1 + 6$) $\times$ Conv($3\times3$, stride 1, 2) \\
Bottleneck size & -- & $64\times19\times19$ \\
Output channels & 2 (add / remove) & 2 (add / remove) \\
Output normalization & Sigmoid & Sigmoid \\
Data augmentation & Rotation ($0$, $90$, $180$, $270^\circ$), flipping & Scaling, rotation ($0$–$360^\circ$), flipping \\
Loss function & BCE & BCE with class weighting \\
Positive class weight & -- & $13\times13\times2$ \\
Learning rate & $1\times10^{-3}$ & $1\times10^{-3}$ \\
Batch size & 128 & 128 \\
Epochs & 200 & 5000 \\
Early stopping & Validation BCE & Validation BCE \\
Hardware & M4 Pro CPU (0.25s/epoch) & RTX 4090 (3s/epoch) \\
\bottomrule
\end{tabular}
\end{table}

\paragraph{Implementation Notes.}
Both versions of NHPM are implemented in PyTorch. Inputs are normalized to the range~$[0,1]$ prior to training. For the discrete maze setting, we perform data augmentation by randomly rotating and flipping the mazes. During continuous-space experiments, we reuse the crowd-sourced mapping data to train the high resolution perception model. To achieve this, we randomly generate terrain maps from maze layouts by performing random scaling, rotation, and flipping. The encoder--decoder model reconstructs the probability maps at the same spatial resolution as the input, ensuring dense, spatially consistent outputs.

\paragraph{Computation Resources}
\label{appendix:computation}
The CNN-based NHPM was trained on an Apple MacBook Pro with an M4 Pro chip using CPU only. Training is efficient, with each epoch taking approximately 0.25 seconds. The mapping task for data collection was deployed on Google Cloud, utilizing a single NVIDIA T4 GPU exclusively for MuJoCo-based visual rendering. The IG-MCTS planning algorithm is lightweight and runs in real time on CPU, without requiring hardware acceleration. To train the bigger neural network for the high resolution perception model for continuous environments, we use a single NVIDIA RTX 4090 GPU. Due to larger input/output and more parameters, each training epoch takes about 3 seconds.

\newpage
\section{Task Reward Composition}
\label{appendix:task_reward}
The \textit{task reward} $R_{\mathrm{task}}$ in our implementation encourages efficient and guided navigation. It is composed of three terms:

\begin{equation}
    R_{\mathrm{task}}(\tau, \zeta) = R_{\mathrm{env}}(\tau) + R_{\mathrm{guidance}}(\tau, \zeta) + R_{\mathrm{smooth}}(\tau),
\end{equation}

where $\tau = (s_1, s_2, \ldots, s_n)$ denotes the sequence of robot states, and $\zeta$ represents the human-suggested path. Each reward component is defined as follows:

\textbf{1. Environment reward.} The robot receives a fixed reward $r_g > 0$ upon reaching a goal location for the first time. Otherwise, it incurs a step penalty of $-1$, encouraging fast progress:

\begin{equation}
R_{\mathrm{env}}(\tau) = 
\begin{cases} 
r_g, & s_n \in \mathcal{G}, \\
-1, & \text{otherwise}.
\end{cases}
\end{equation}

\textbf{2. Guidance reward.} To align with human intent, the robot is penalized for deviating from the suggested path $\zeta$. The penalty increases with the number of steps since the last interaction:
\begin{equation}
R_{\mathrm{guidance}}(\tau, \zeta) = 
\begin{cases} 
0, & s_n \in \zeta, \\
-\log(n), & \text{otherwise}.
\end{cases}
\end{equation}

Intuitively, $R_{\mathrm{guidance}}$ is zero if the robot's current state lies on the human-provided path. Otherwise, the penalty grows logarithmically with the episode length $n$, encouraging the robot to either stay on track or communicate again before straying too far.

\textbf{3. Smoothness reward.} This term penalizes revisiting previously visited states, promoting smooth and efficient trajectories:

\begin{equation}
R_{\mathrm{smooth}}(\tau) = -\sum_{i=1}^{n-1} \mathbbm{1}[s_i = s_n].
\end{equation}

\paragraph{Implementation-specific design details.} 
To reduce the search horizon, lower estimation variance, and improve computational efficiency, we terminate an MCTS rollout early if the agent either (a) reaches the goal or (b) performs a communication action. However, this early stopping introduces a bias toward shorter trajectories. To counteract this, we impose a communication cost of $c = 10 + n$, where $n$ denotes the number of human-suggested states in $\zeta$ that the robot has not yet visited. This discourages premature communication and better aligns with long-term task performance.

\newpage
\section{Algorithm Pseudocode}\label{appendix:pseudocode}
\begin{algorithm}[htbp]
\caption{IG-MCTS}
\label{algo:IG_MCTS}
\begin{algorithmic}[1]
    \STATE \textbf{Input:} human guidance $\zeta$, previous state-visitation history $\tau_0$, current human perception state $x_0$
    \STATE \textbf{Parameters:} iterations $n=100$, exploration constant $k=\sqrt{2}$, discount factor $\gamma=0.99$, depth $d=100$
    \STATE Create root node $v_0$ with $(\tau_0, x_0)$, initialize $Q(v_0) \gets 0$, $N(v_0) \gets 0$, $\mathbb{C}(v_0) \gets \emptyset$
    \FOR{each iteration $i$ from 1 to $n$}
        \STATE Set $v \gets v_0$, $\mathrm{stopping} \gets \text{False}$
        \WHILE{$v$ is not terminal and $\mathrm{stopping} = \text{False}$}
            \IF{$v$ is fully expanded}
                \STATE $v \gets \arg\max_{v' \in \mathbb{C}(v)} \left( \frac{Q(v')}{N(v')} + k \sqrt{\frac{\log N(v)}{N(v')}} \right)$
            \ELSE
                \STATE $v, \mathrm{stopping} \gets \textsc{Expand}(v)$
            \ENDIF
        \ENDWHILE
        \STATE $q \gets \textsc{Rollout}(v)$
        \STATE \textsc{BackPropagate}($v, q$)
    \ENDFOR
    \STATE \textbf{Return} action of best child $c^\star = \arg\max_{c \in \mathbb{C}(v_0)} N(c)$
\end{algorithmic}
\end{algorithm}

\begin{minipage}{0.33\textwidth}
    \begin{algorithm}[H]
        \captionsetup{labelformat=default,labelsep=colon,name=Procedure}
        \caption{\sc Expand($v$)}
        \label{algo:expand}
            \begin{algorithmic}[1]
            \STATE Choose an untried action $a\in \Acal$
            \IF{$a\in \Ucal$}
                \STATE $s=\mathrm{last}(\tau)$
                \STATE $\tau'\gets \tau\oplus T(s,a)$
                \STATE Create node $v'$ with $(\tau', x)$
                \STATE $\mathbb{C}(v) \gets \mathbb{C}(v) \cup \{v'\}$
                \STATE \textbf{Return} $v'$, False
            \ELSIF{$a\in \Ocal$}
                \STATE $x'\gets F(x, (\tau,a))$
                \STATE Create node $v'$ with $(\tau, x')$
                \STATE \textbf{Return} $v'$, True
            \ENDIF
            \end{algorithmic}
    \end{algorithm}
\end{minipage}
\begin{minipage}{0.33\textwidth}
    \begin{algorithm}[H]
        \captionsetup{labelformat=default,labelsep=colon,name=Procedure}
        \caption{\sc Rollout($v$)}
        \label{algo:rollout}
        \begin{algorithmic}[1]
            \STATE Initialize $q \gets 0$, depth $d \gets 0$
            \WHILE{$d < T$ and $(\tau, x)$ not terminal}
                \STATE Sample $a \in A$
                \IF{$a\in U$}
                    \STATE $\tau\gets \tau\oplus T(s,a)$ where $s=\mathrm{last}(\tau)$
                \ELSIF{$a\in O$}
                    \STATE $x'\gets F(x, (\tau,a))$
                \ENDIF
                \STATE $q \gets q + \gamma^d R(\tau, \zeta, x, x')$
                \STATE $x\gets x'$
                \STATE $d \gets d + 1$
            \ENDWHILE
            \STATE \textbf{Return} $q$
        \end{algorithmic}
    \end{algorithm}
\end{minipage}
\begin{minipage}{0.33\textwidth}
    \centering
    \begin{algorithm}[H]
        \captionsetup{labelformat=default,labelsep=colon,name=Procedure}
        \caption{\sc BackPropagate($v, q$)}
        \label{algo:backprop}
        \begin{algorithmic}[1]
            \STATE Initialize $q_{\text{sample}} = q$ and $\delta=1$
            \WHILE{$v$ is not null}
                \STATE Current value estimate $w = \frac{Q(v)}{N(v)}$ if $N(v)>0$ else $0$
                \STATE $q_{\text{sample}} \gets r(v) + \gamma\left[\delta q_{\text{sample}} + (1-\delta) w\right]$
                \STATE $Q(v) \gets Q(v) + q$
                \STATE $N(v) \gets N(v) + 1$
                \STATE $\delta\gets \delta(v)$
                \STATE $v \gets$ parent of $v$
            \ENDWHILE
        \end{algorithmic}
    \end{algorithm}
\end{minipage}

\section{Baseline Pupil Diameter Measurement}\label{appendix:baseline_pupil}
To account for individual differences in pupil sizes, we ask each participant to read a brief text paragraph at the beginning of each session. The eye-tracking data from this period is used to calculate the mean pupil diameter as the baseline for the session's PCPD. We drafted these paragraphs to ensure comparable length and maintain neutral content.

\begin{small}
\begin{tcolorbox}[colback=gray!5, colframe=black!80, title=Text Before Method A (Teleoperation)]
Making a sandwich begins by picking your favorite type of bread. You can spread butter, mayonnaise, or other condiments before adding a layer of vegetables, meat, or cheese. Once the ingredients are in place, press the slices together gently. Preparing a sandwich is a simple task, but it’s also a quick and satisfying way to create a meal.
\end{tcolorbox}
\begin{tcolorbox}[colback=gray!5, colframe=black!80, title=Text Before Method B (Instruction-Following)]
Washing dishes starts by filling the sink with warm, soapy water. Plates, bowls, and utensils are scrubbed clean with a sponge to remove food residue. Once clean, they are rinsed under running water and placed on a rack to dry. While it’s a routine chore, it’s also a small step toward keeping the kitchen tidy and organized.
\end{tcolorbox}
\begin{tcolorbox}[colback=gray!5, colframe=black!80, title=Text Before Method C (IG-MCTS)]
Sitting in a chair can be a relaxing moment during a busy day. You adjust your position to get comfortable, letting your body rest as you settle in. Sometimes, it’s a chance to pause and think quietly. Whether you’re sitting to read, work, or simply take a break, it’s a small but familiar part of daily life.
\end{tcolorbox}
\end{small}

\subsection{PCPD Calculation}\label{appendix:pcpd_calculation}
The percent change in pupil diameter (PCPD) is calculated as:
\begin{equation}
    \text{PCPD}_t = \frac{\text{pupil diameter}_t - \text{baseline diameter}}{\text{baseline diameter}}
\end{equation}
Here, \( \text{baseline diameter} \) is the average pupil diameter recorded during the baseline period, as detailed in \Cref{appendix:baseline_pupil}. For each recording, we compute the mean and standard deviation of PCPD over time. These statistics are then aggregated, and we report the averages in \Cref{tab:eye_metrics}.

\section{Maze Layouts}\label{appendix:maze_layouts}
The three maze layouts used in the user study were generated from fixed seeds on a Linux system (MuJoCo's randomization differs across operating systems). The layouts are visualized below:

\begin{minipage}{0.32\textwidth}
    \begin{tcolorbox}[colback=gray!5, colframe=black!80, boxrule=0.5mm, 
        sharp corners,  
        title={
            \centering
            \textbf{Layout 1} \\
            \small Seed: 234 \\
        }, width=\textwidth]
        \centering
        \includegraphics[width=\textwidth]{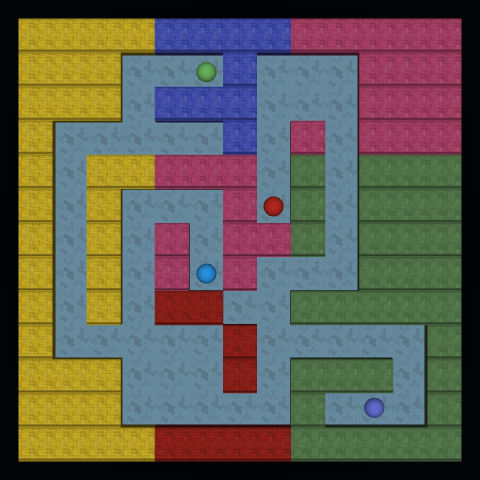}
        \\
        \small Robot Groundtruth
        \\[1em]
        \includegraphics[width=\textwidth]{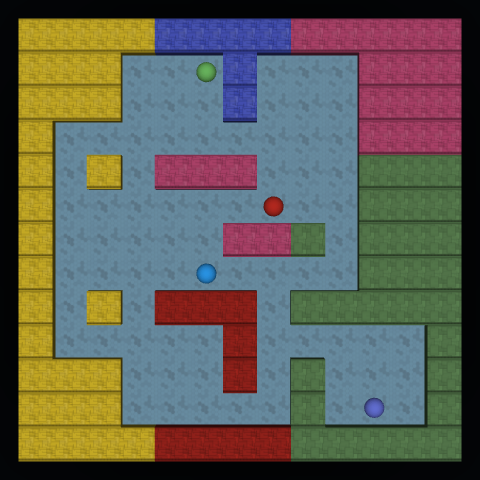}
        \\
        \small Initial Human Map
    \end{tcolorbox}
\end{minipage}
\hfill
\begin{minipage}{0.32\textwidth}
    \begin{tcolorbox}[colback=gray!5, colframe=black!80, boxrule=0.5mm, 
        sharp corners,
        title={
            \centering
            \textbf{Layout 2} \\
            \small Seed: 666 \\
        },
        width=\textwidth]
        \centering
        \includegraphics[width=\textwidth]{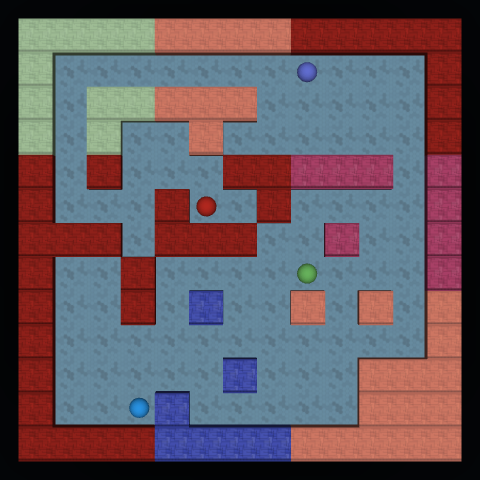}
        \\
        \small Robot Groundtruth
        \\[1em]
        \includegraphics[width=\textwidth]{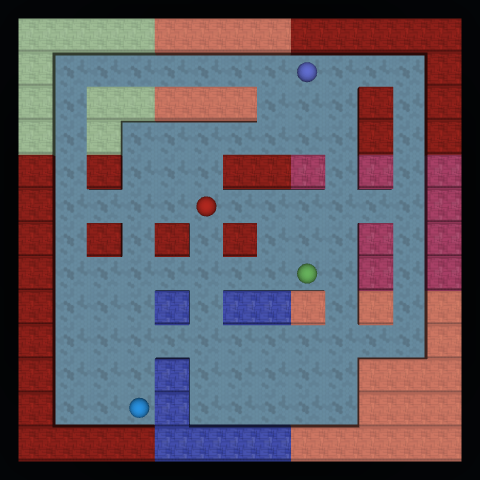}
        \\
        \small Initial Human Map
    \end{tcolorbox}
\end{minipage}
\hfill
\begin{minipage}{0.32\textwidth}
    \begin{tcolorbox}[colback=gray!5, colframe=black!80, boxrule=0.5mm, 
    sharp corners,
    title={
            \centering
            \textbf{Layout 3} \\
            \small Seed: 9 \\
        },
        width=\textwidth]
        \centering
        \includegraphics[width=\textwidth]{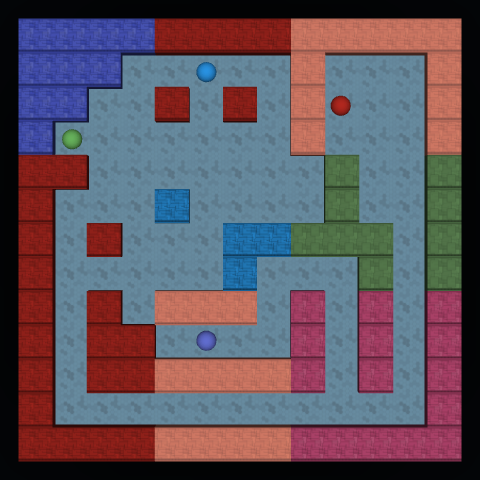}
        \\
        \small Robot Groundtruth
        \\[1em]
        \includegraphics[width=\textwidth]{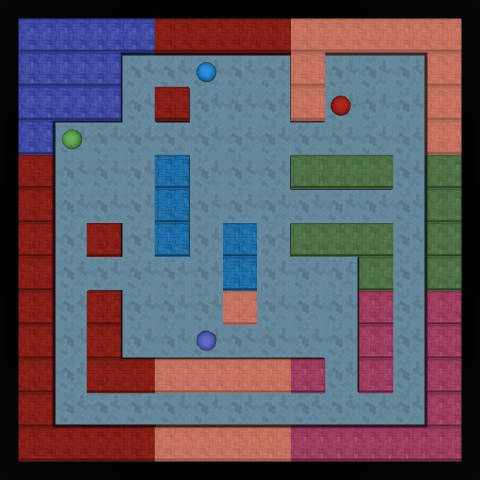}
        \\
        \small Initial Human Map
    \end{tcolorbox}
\end{minipage}


\end{document}